\documentclass[runningheads]{llncs}

\def\BibTeX{{\rm B\kern-.05em{\sc i\kern-.025em b}\kern-.08em
    T\kern-.1667em\lower.7ex\hbox{E}\kern-.125emX}}

\usepackage{cite}
\usepackage{amsmath,amssymb,amsfonts}
\usepackage{algorithmic}
\usepackage{graphicx}
\usepackage{textcomp}
\usepackage{xcolor} 
\usepackage{url}
\usepackage{xspace}
\usepackage{tikz}
\usetikzlibrary{arrows,shapes}

\usetikzlibrary{decorations.pathreplacing}
\makeatletter
\def\underbrace#1{%
	\@ifnextchar_{\tikz@@underbrace{#1}}{\tikz@@underbrace{#1}_{}}}
	\def\tikz@@underbrace#1_#2{%
		\tikz[baseline=(a.base)] {\node[inner sep=4] (a) {\(#1\)};
			\draw[line cap=round,decorate,decoration={brace,amplitude=5pt}]
			(a.south east) -- node[below,inner sep=4pt] {\(\scriptstyle #2\)} (a.south west);}}

\newcommand{\ourtool}{\textsc{SMarTplan}\xspace}
\newcommand{\omtplan}{RCLLPlan\xspace}
\newcommand{\lcp}{LCP\xspace}
\newcommand{\eg}{\textit{e.g.}\xspace} 
\newcommand{\ie}{\textit{i.e.}\xspace}

\begin{document}

\title{\ourtool{}: a Task Planner for Smart Factories
\thanks{The research of Arthur Bit-Monnot and Luca Pulina has been funded by the EU Commission’s H2020 Programme under grant agreement N.
	732105 (CERBERO project). The research of Luca Pulina was also partially supported by a grant from Regione Sardegna (Italy), {\it POR FESR 2014/2020 -  Asse Prioritario I "Ricerca Scientifica, Sviluppo Tecnologico e Innovazione"} (PROSSIMO project).}
}
  
\author{Arthur Bit-Monnot\inst{1} \and Francesco Leofante\inst{2,3} \and Luca Pulina\inst{1} \and Erika \'Abrah\'am\inst{3} \and Armando Tacchella\inst{2}}

\institute{University of Sassari, Sassari, Italy \\ \email{\{afbit,lpulina\}@uniss.it} \and RWTH Aachen University, Germany\\
 \email{\{leofante,abraham\}@cs.rwth-aachen.de} \and University of Genoa, Genoa, Italy\\ \email{armando.tacchella@unige.it}  }

\authorrunning{A. Bit-Monnot et al.}

\maketitle
\thispagestyle{plain}
\pagestyle{plain}

\begin{abstract}
  Smart factories are on the verge of becoming the new industrial paradigm, wherein optimization permeates all aspects of production, from concept generation to sales.
  To fully pursue this paradigm, flexibility in the production means as well as in their timely organization is of paramount importance.
  AI is planning a major role in this transition, but the scenarios encountered in practice might be challenging for current tools.
  Task planning is one example where AI enables more efficient and flexible operation through an online automated adaptation and rescheduling of the activities to cope with new operational constraints and demands.
  In this paper we present \ourtool{}, a task planner specifically conceived to deal with real-world scenarios in the emerging smart factory paradigm.
  Including both special-purpose and general-purpose algorithms, \ourtool{} is based on current automated reasoning technology and it is designed to tackle complex application domains. In particular, we show its effectiveness on a logistic scenario, by comparing its specialized version with the general purpose one, and extending the comparison to other state-of-the-art task planners.
\end{abstract}

\section{Introduction}
\label{sec:intro}
In recent years manufacturing is experiencing a major paradigm shift due to a
variety of drivers. From the market side, the push towards 
high product customization led to a higher proliferation of
product variants. From the organizational side, the need to take into
account customer feedback led to shorter product cycles. From the
technological side, the convergence between traditional industrial
automation and information technology brought additional opportunities
by combining fields such as, \eg, intelligent Cyber-Physical Systems, additive manufacturing, cloud
computing, and the Internet of Things -- see, \eg,~\cite{zhong2017intelligent} for a recent survey.

The short planning horizons and
product life cycles induce the decrease of batch sizes and do
therefore require manufacturing flexibility.
In order to make the right management decisions, real-time
information and the direct realization of the decisions are
mandatory. The term \textit{smart factory} is often used to refer to
the combination of industrial automation and information technology
that should respond to the change drivers, and become the prevalent
industrial paradigm, wherein optimization permeates all aspects of
production, from concept generation to sales.

To fully pursue this paradigm, flexibility in production processes as
well as in their timely organization is of paramount importance.
AI planning has the potential to play a major role in this transition, allowing for more efficient and flexible operation through an on-line automated adaptation and rescheduling of the activities to cope with new operational constraints and demands.

However, the scenarios
encountered in practice might be challenging for current
state-of-the-art tools. For instance, task planning for production logistics 
often requires reasoning on expressive models that combine ordering
constraints,  time windows and spatial relations which most tools
cannot handle and solve efficiently. Furthermore, task planning for
high product variety may pose additional challenges to planners that
rely on grounding procedures. Even if solving the problem might
actually be easy, instantiation of the entire planning task may swamp
the memory and hamper the search for a feasible solution.   

With the above considerations in mind, in this paper we
present \ourtool{}, a task planner specifically 
conceived to deal with real-world scenarios in the emerging smart
factory paradigm. Including both special-purpose and general-purpose
algorithms, \ourtool{} is based on current automated reasoning
technology, namely Satisfiability Modulo Theories (SMT) and
Optimization Modulo Theories (OMT) solving ---
see, \eg, \cite{dblp:series/faia/barrettsst09,dblp:conf/tacas/cimattifgss10}
and \cite{z3,dblp:conf/tacas/bjornerpf15} for related solvers. SMT
solvers developed into a crucial technology in many areas of computer-aided verification ---
see, \eg, \cite{DBLP:conf/cav/CimattiG12,DBLP:conf/cade/TiwariGD15}; their application has also
been explored in task planning~\cite{smtplan,bit-monnot-cp-2018}. OMT solvers have been introduced more recently, but they
already showed promise in some tasks, including task
planning~\cite{dblp:conf/iri/leofanteanlt17,dblp:journals/isf/leofanteanlt18}.
The combination of leading-edge SMT and OMT decision procedures with 
effective encodings, is our recipe to tackle relevant application domains. 

In particular, we show the effectiveness of \ourtool{} on a logistic
scenario based on the  RoboCup Logistics League
(RCLL)~\cite{PlanningInRCLL}, wherein two teams of autonomous robots
compete to handle the  logistics of materials through several dynamic
stages to  manufacture products in a smart factory scenario. Using the
RCLL as a  testbed, we aim to \textit{(i)} compare the performances
of \ourtool{} with other general purpose planners and \textit{(ii)}
compare the performances of its specialized version 
with the general purpose one to highlight strengths and weaknesses of
each. The main contribution of our paper is to show that, while
considerable investment in research and tech-transfer is required to
cope with smart factory needs as a whole, task planners can be made
fit to leverage either domain knowledge or suitable
encodings towards state-of-the-art decision procedures.

The remainder of this paper is structured as
follows. Section~\ref{sec:background} introduces background knowledge
on which \ourtool{} builds, while Section~\ref{sec:architecture} gives
an overview on its architecture and internals. The logistic scenario
used to assess the performance of our tool is presented in
Section~\ref{sec:rcll}, followed by Section~\ref{sec:experiments}
where experimental results are discussed. Concluding remarks together with
future directions are discussed in Section~\ref{sec:conclusion}.  

\section{Background}
\label{sec:background}

\subsection{Task Planning}

Task planning is a field of Artificial Intelligence concerned with finding a set of actions that would result in desirable state.
It is traditionally formulated as a state transition system in which actions allow to transition from one state to another. 
Solving a planning problem means finding a sequence of actions, i.e. a path, from an initial state to a goal state.

Most research in automated planning has focused in heuristic search techniques in which forward search planners explore the set of states that are reachable from the initial one.
Such techniques have proved to very efficiently handle large problems in classical planning, through development of targeted heuristic functions.
Such techniques have however proved to be difficult to extend to richer problems such as those involving a rich temporal representation or continuous variables, leading to a renewed interest in constraint-based approach (e.g. \cite{bit-monnot2016,smtplan}).

\subsection{Planning as Satisfiability}

As shown first shown in~\cite{DBLP:conf/ecai/KautzS92}, classical planning
problems can be naturally formulated as propositional satisfiability
problems and solved efficiently by SAT solvers. The idea is to encode
the existence of a plan of a fixed length $p$ as the satisfiability of a propositional logic formula: the formula
for a given $p$ is satisfiable if and only if there is a plan of
length $p$ leading from the initial state to the goal state, and a
model for the formula represents such plan.

Classical planning abstracts aways from time and assumes the actions and state changes to be instantenous. In contrast, \textit{temporal planning} considers action durations and temporal relations between (possibly concurrently executed) actions. For instance, in logistics plans might need to meet deadlines in order to satisfy some production requirements. The natural encoding of temporal planning problems requires an extension of propositional logic with arithmetic theories, such as the theory of reals or integers. Recent advances in satisfiability checking \cite{dblp:series/faia/barrettsst09} led to powerful \emph{Satisfiability Modulo Theories} (\emph{SMT}) solvers such as \cite{DBLP:conf/tacas/CimattiGSS13,DBLP:conf/sat/CorziliusKJSA15,z3}, which can be used to check the satisfiability of first-order logic formulas over arithmetic theories and thus to solve temporal planning problems.

\subsection{SMT and optimization}

Satisfiability Modulo Theories is the problem of deciding the
satisfiability of a first-order formula with respect to some decidable
theory $\mathcal{T}$. In particular, SMT generalizes the boolean
satisfiability problem (SAT) by adding background theories such as the
theory of real numbers, the theory of integers, and the theories of
data structures (\textit{e.g.}, lists, arrays and bit vectors).

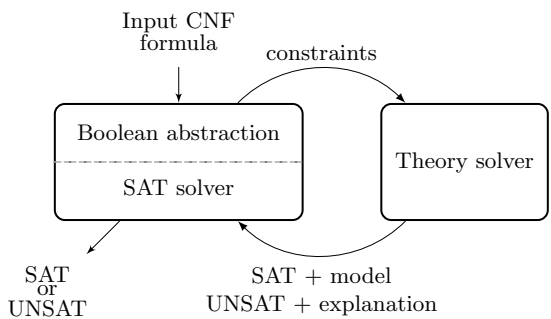
\begin{figure}[b]
	\begin{center}
	  \scalebox{0.95}{

\begin{tikzpicture}[thick]

\pgfdeclarelayer{background}
\pgfdeclarelayer{foreground}
\pgfsetlayers{background,main,foreground}

\tikzstyle{doc}=[%
draw,
thick,
align=center,
color=black,
shape=document,
minimum width=10mm,
minimum height=15.2mm,
shape=document,
inner sep=2ex,
]

    \node [rectangle, draw,text centered, rounded corners, text width=10em, minimum height=5em, minimum width = 10em] (sat) {    	
    	\begin{tabular}{c}
    	Boolean abstraction\\
    	\\
    	SAT solver
    	\end{tabular}};

    \node[above of=sat, node distance=1.8cm](cnf){\begin{tabular}{c}
    	Input CNF\\[-0.08cm]
    	formula
    	\end{tabular}};
    
    \begin{pgfonlayer}{foreground}
    \path (sat.west |- sat.west) node (a) {};
    \path (sat.east -| sat.east)node (b) {};
    \path[ draw=black!50, dashed]
    (a) rectangle (b);
    \end{pgfonlayer}
    
    \node [rectangle, draw,text centered, rounded corners, thick, text width=6.5em, minimum height=5em, minimum width =6.5em, right of=sat, node distance=4cm] (smt) {Theory solver};

	\node[below left of=sat, node distance=2.6cm](res){\begin{tabular}{c}
		SAT \\[-0.2cm]
		or\\[-0.1cm]
		UNSAT
		\end{tabular}};

	\path [draw, -latex'] (cnf) -- (sat);
	\draw[-latex,bend left=45]  (sat) edge [above] node {constraints} (smt);
	\draw[-latex,bend right=-45]  (smt) edge  [below] node {\begin{tabular}{c}
		SAT + model \\
		UNSAT + explanation
		\end{tabular}} (sat);
	\path [draw, -latex'] (sat) -- (res);
	
\end{tikzpicture}
}
	\end{center}
	\vspace*{-2ex}
	\caption{The SMT solving framework.}
	\label{fig:smt}
\end{figure}

To decide the satisfiability of an input formula $\varphi$ in Conjunctive Normal Form (CNF), SMT solvers (see Fig. \ref{fig:smt}) typically first build a \emph{Boolean abstraction} $\textit{abs}(\varphi)$ of $\varphi$ by replacing each constraint by a fresh Boolean variable (proposition), \eg,

\begin{eqnarray*}
	\arraycolsep=2pt
	\begin{array}{ccccccccccc}
		\varphi &: &\underbrace{x \geq y} &\wedge &(&\underbrace{y > 0}& \vee &\underbrace{x >0}&)& \wedge &\underbrace{y \leq 0} \\
		\textit{abs}(\varphi)&:&A& \wedge& (&B& \vee &C&)& \wedge  &\neg B
	\end{array}
	\label{eq:abs}
\end{eqnarray*}
where $x$ and $y$ are real-valued variables, and $A$, $B$ and $C$ are propositions.

\noindent A SAT solver searches for a
satisfying assignment $S$ for $\textit{abs}(\varphi)$, \eg, $S(A)=\top$, $S(B)=\bot$,
$S(C)=\top$ for the above example.  If no such assignment exists then the
input formula $\varphi$ is unsatisfiable. Otherwise, the
consistency of the assignment in the underlying theory
is checked by a \emph{theory solver}. In our example, we check whether the set $\{ x \geq y,\ y
\leq 0,\ x > 0\}$ of linear inequalities is feasible, which is the
case. If the constraints are consistent then a satisfying solution
(\textit{model}) is found for $\varphi$. Otherwise, the theory solver returns a theory lemma $\varphi_E$ giving an
\textit{explanation} for the conflict, \eg, the negated conjunction of some inconsistent input constraints.
The explanation is used to refine the Boolean abstraction $\textit{abs}(\varphi)$ to $\textit{abs}(\varphi)\wedge \textit{abs}(\varphi_E)$.
These steps are iteratively executed until either a theory-consistent
Boolean assignment is found, or no more Boolean satisfying assignments
exist.

Standard decision procedures for SMT have been extended with optimization capabilities, leading to Optimization Modulo Theories (OMT). OMT extends SMT solving with optimization procedures to find a variable assignment that 
defines an optimal value for an objective function $f$ (or a combination of multiple
objective functions) under all models of a formula $\varphi$. As noted in~\cite{DBLP:journals/tocl/SebastianiT15}, OMT solvers such as~\cite{DBLP:conf/cav/SebastianiT15,dblp:conf/tacas/bjornerpf15} typically implement a \textit{linear search} scheme, which can be summarized as follows.
Let $\varphi_S$ be the conjunction of all theory constraints that
are true under $S$ and the negation of those that are false under $S$. A local optimum $\mu$ for $f$ is computed\footnote{For instance, if $f$ and $\varphi_S$ are expressed in  \texttt{QF\_LRA}, this can be done with Simplex} under the
side condition $\varphi_S$ and $\varphi$ is updated as 

\begin{equation*}
\label{eq:optimization}
\varphi := \varphi \wedge (f \bowtie \mu) \wedge \neg \bigwedge \varphi_S  \quad  ,\quad \bowtie \in \{<, >\} 
\end{equation*}

\noindent Repeating this procedure until the formula becomes unsatisfiable will lead to an assignment minimizing $f$ under all models of $\varphi$.

\section{The architecture of \ourtool{}}
\label{sec:architecture}

\ourtool is a task planner that leverages the latest SMT and OMT technology to solve smart factory problems.
It comes with two planning components:
\begin{itemize}
  \item \omtplan  is specifically tailored for the RCLL problems.
    Given problem desciption in PDDL format, it generates domain specific encodings for both SMT and OMT, thus supporting both satisfiability as well as optimal planning.
  \item \lcp is the domain-independent component and accepts as input arbitrary PDDL domain and problem files.
  \lcp internally generates time-oriented SMT encoding which are solved using an off-the-shelf SMT solver.
\end{itemize} 

For both planning components, the resulting plan is extracted from the SMT/OMT encoding and validated against VAL \cite{DBLP:conf/ictai/HoweyLF04}.

\begin{figure}[t]
\begin{center}
  \scalebox{0.8}{

\tikzstyle{database} = [cylinder,
	cylinder uses custom fill,
	cylinder body fill=blue!10,
	cylinder end fill=blue!10,
	shape border rotate=90,
	aspect=0.18,
	minimum height=1.5cm,
	draw]

	\begin{tikzpicture}[thick]
		\node[database]  (database) {\begin{tabular}{c}
			Problem\\
			data
			\end{tabular}};

		\node[draw, rounded corners, right of= database, node distance = 5cm, minimum width=6cm, minimum height=8cm,  align=left] (planner) {};
		
		\node[above, xshift=3cm,yshift=-1cm] at (planner.north west) {\textbf{\ourtool{}}};
		
		\node[below, draw, rounded corners, minimum height=3cm, minimum width=3cm,yshift=-0.5cm] at (planner.base) (dahu)  {};
		
		\node[above, xshift=1.4cm, yshift=1cm] at (dahu.base west) {LCP};
		
		\node[below, draw, rounded corners, minimum height=0.8cm, minimum width=0.7cm, xshift=0.8cm, yshift=-0.5cm] at(dahu.base west) (pddl) {PDDL};
		
		\node[above of=pddl](plus) {$\bigoplus$};
	
		\node[above right of=pddl, draw, rounded corners, minimum height=0.8cm, minimum width=1cm,node distance=1.5cm, xshift=0.3cm, yshift=-0.05cm] (smt) {SMT};

		\node[above, draw, rounded corners, minimum height=2cm, minimum width=3cm,yshift=0.5cm] at (planner.base) (spec)  {};
		
		\node[above, xshift=1.5cm, yshift=0.5cm] at (spec.base west) {\omtplan};
		
		\node[below, draw, rounded corners, minimum height=0.8cm, minimum width=0.7cm, xshift=1.5cm, yshift=0cm] at(spec.base west) (omt) {SMT/OMT};
		
		\node[draw, rounded corners, right of=database, node distance=7.5cm, rotate=90, minimum width=5cm](pp) {Pretty Printer \& Validation};
		
		\node[right of=pp,node distance=2cm](plan) {Plan};

		\path [draw, -latex'] (pddl) -- (plus.south);
		\path [draw, -latex'] (plus) -- (smt);
		\path [draw, -latex'] (database) -- ++ (2.5,0) -- ++(0,-1.92) -- (plus);
		
		\path [draw, -latex'] (database) -- ++ (2.5,0) -- ++(0,1.1) -- (omt);
		
		\path [draw, -latex'] (smt.east) -- ++(0.7,0) -- ++(0,1.92) -- (pp);
		\path [draw, -latex'] (omt.east) -- ++(0.945,0) -- ++(0,-1.1) -- (pp);
		\path [draw, -latex'] (pp) -- (plan);

	\end{tikzpicture}

  }
	\caption{Architecture of \ourtool.}
	\label{fig:architecture}
\end{center}
\end{figure}

\subsection{Special purpose solution}
\label{sub:special}
\omtplan implements domain-specific encodings of the \textit{state-based} planning problem defined over the scenario we target here. This module builds upon previous work~\cite{dblp:conf/iri/leofanteanlt17,leofanteanlt2018} and extends it with the ability to generate different types of encodings specifically tailored to be able to cope with the complexity of the domain. In particular, we present here \textit{(i)} a fine-grained encoding where single actions are encoded separately and \textit{(ii)} an encoding which leverages domain-specific knowledge to build more compact encodings that enable macro planning by grouping action constraints together. In both cases, \omtplan can leverage SMT or OMT technology to produce either \textit{feasible} or \textit{optimal} plans, solving a planning as satisfiability problem defined as follows.

World states are described using an ordered set of real-valued \emph{variables} $x = \{ x_1,\ldots,x_n\}$. We also use the vector notation $x = (x_1,\ldots,x_n)$ and write $x'$ and $x_i$ for $(x_1',\ldots,x_n')$ and $(x_{1,i},\ldots,x_{n,i})$ respectively. We use special variables $A\in x$ to encode the \emph{action} to be executed at each step and $t  \in x$ for the associated time stamp. A \emph{state} $s = (v_1,\ldots,v_n)\in\mathbb{R}^n$ specifies a real value $v_i \in \mathbb{R}$ for each variable $x_i\in x$. 

The RCLL domain is represented symbolically by mixed-integer arithmetic formulas defining the \emph{initial states} $I(x)$, the \emph{transition relation} $T(x,x')$ (where $x$ describes the state before the transition and $x'$ the state after it) and a set of \emph{final} states $F(x)$. The transition relation is defined in terms of actions that can be performed at each step.
A \emph{plan} of length $p$ is a sequence $s_0,\ldots,s_p$ of states such that $I(s_0)$ and $T(s_i,s_{i+1})$ hold for all $i=0,\ldots,p-1$, and $F(s_p)$ holds. Thus, plans are models for the formula:

\vspace*{-2ex}
{
	\begin{eqnarray}
	\label{eq:BSR}
	 I(x_0) \wedge \left( \bigwedge_{0\leq i < p} T(x_i,x_{i+1}) \right) \wedge \left(\bigvee_{0\leq i \leq p} F(x_i) \right) 
	\end{eqnarray}
}

In general the length of a plan is not known a priori and has to be determined empirically by increasing $p$  until a satisfying assignment for \ref{eq:BSR} is found. However, \omtplan, being tailored for the domain considered, exploits domain specific knowledge to determine an upper bound on $p$ and simplify plan search.
In order to be able to support generation of optimal plans with OMT, the bounded planning approach is extended to enable optimization over cost structures expressed in first-order arithmetic theories. We introduce additional variables $\textit{c}\in x$ to encode the \emph{cost} of executing action $A$ at time $t$. We define the total cost $\textit{c}_{\textit{tot}}$ associated to a plan as:

\vspace*{-2ex}
{
	\begin{eqnarray}
	\label{eq:objective}
	&& \textit{c}_{\textit{tot}} = \sum_{0 \leq i < p} \textit{c}_i
	\end{eqnarray}
}

\emph{Optimal bounded planning} is then defined as the problem to find a path of length at most $p$ that reaches a target state and achieves thereby the smallest possible cost, \ie, to minimize Eq.~\ref{eq:objective} under the side condition that Eq.~\ref{eq:BSR} holds.\footnote{\omtplan exploits a simple cost definition in its current state, \ie, minimize time to delivery for each product. However, richer goal structures could be specified.}

The same ideas presented above are applied when generating encodings for macro actions. In this case, domain-specific knowledge is encoded explicitly at the logical level, so as to produce encodings that are more compact and easier to solve. Macro actions are encoded based on the following observation. Plans for production in the RCLL often involve action sequences that yield better performance if performed by the same agent. For instance, if a robot is instructed to prepare a base station to provide a base, then it makes sense that the same robot also retrieves the base (under the realistic assumption that in the RCLL providing a base is less expensive than motion planning and navigation for another robot). Following such observations, a logical encoding is built in \omtplan  where the transition relation contains constraints for macro actions only.

\subsection{General purpose algorithm}
\label{sub:general}

This domain-dependent planner is complemented by LCP (Lifted Constraint Planner) a domain-independent planner that aims at providing good performance over a large set of problem.
LCP supports both temporal PDDL~\cite{Fox2003} and a subset of ANML~\cite{anml2008} to define planning problems.

Like \omtplan, it uses a constraint-based encoding where multi-valued variables are related through a set of constraints that can be exploited by SMT solvers.
Unlike \omtplan's state-oriented representation, LCP relies on time-oriented encoding where the effects and conditions of actions are placed on temporal intervals which are related through temporal constraints ensuring the consistency of a plan.

\subsubsection{From PDDL and ANML to Chronicles}

LCP uses chronicles~\cite{ghallab2004} as a building block for its internal representation.
Chronicles were first introduced in the IxTeT planner \cite{ghallab1994}. 
They allow an expressive representation of temporal actions that leverages a constraint-based representation close to the one found in state-of-the-art scheduling solvers such as CP Optimizer \cite{laborie2008}.

The environment is represented by a finite set of state variables, that describe the evolution of a particular state feature overtime, e.g., the location of a robot over the course of the plan.

A chronicle is composed of a set of decision variables, a set of constraints relating those variables as well as some conditions and effect statements.
Condition statements require a particular state variable to have a given value over a temporal interval while effect statements change the value of a state variable at given point in time.
In essence, a chronicle is thus a CSP extended with additional constructs to represent the conditions and effects that are at the core of AI planning.

Chronicles offer a natural representation for the action models present in both the PDDL and  ANML languages that are supported by our tool.
In practice, one simply needs to map an action's parameters and timepoints into to variables of the chronicles. The action's condition and effect can be straightforwardly encoded into the corresponding condition and effect statements.
Additional requirements, e.g. on the duration of an action, are encoded as constraints on the chronicles variables.
For this translation, we follow the process demonstrated by other planners such as IxTeT \cite{ghallab1994} and FAPE \cite{bit-monnot2016}.

The planning problem itself is also encoded as a chronicle, with effect statements defining the initial state and condition statements representing the goals of the problem.

\subsubsection{From Chronicles to Constraint Satisfaction Problems}

Planning differs from scheduling in that the actions that will be part of the solution plan are not known beforehand.
We escape this problem by generating bounded problems in which a finite set of actions are allowed to be part of a solution plan.
More precisely, a bounded planning problems has a finite set of action chronicles, each representing a possible action in the solution plan.
Action chronicles are optional: they are associated to a boolean decision variables that is true if the action is part of the solution plan and false otherwise.

Finding a solution to a bounded planning problem means finding an assignment to decision variables that represent action parameters (\ie variables inside the chronicles) and action presence (the boolean variables) such that the set of actions that are present form a consistent plan.

Plan consistency is defined through a set of constraints over the chronicles. Those constraints are detailed in \cite{bit-monnot-cp-2018} and are sketched below:
\begin{itemize}
    \item \emph{consistency constraints} enforce that no two effects statements are overlapping.
    Namely, if two effect statements are part of actions that are present in the solution, they must either affect different state variables or be active on non-overlapping temporal intervals.
    \item \emph{support constraints} ensure that any condition statement in an action that is part of the solution is supported by an effect statement.
    More precisely, it means that there exists an effect statements that changes the state variable to the value required by the condition.
    In addition, the effect statement must be active before the condition, and they must be no other effect affecting the same state variable active between the start of the effect and the end of the condition.
    \item \emph{internal constraints} ensure that all constraints defined inside a chronicle hold if the corresponding action is part of the solution.
\end{itemize}

For a given bounded planning problem, a CSP is built by taking the conjunction of all consistency, support and internal constraints.

This formulation has some important similarities with lifted plan-space planning \cite{ghallab1994,bit-monnot2016,dvorak-2014-ictai}.
It differs from the former in that the CSPs in LCP contain no dynamic part since possible actions are fixed beforehand. This facilitates the use of off-the-shelf solvers while plan-space planner typically require ad-hoc constraint solving engines.
The representation is also closely related to the one of recent constraint programming engines for scheduling that support of optional temporal intervals but lack a notion of condition and effects \cite{laborie2008,laborie2009}.

\subsubsection{Planning}

Planning is done by generating increasingly large bounded planning problems.
At each step, an SMT solver is used to prove the existence or absence of a plan for a bounded planning problem, \ie, whether a plan exists with the limited number of actions allowed.
The SMT solver is used to find a satisfying assignment (model), which can then be translated into a solution plan. When the solver proves the inconsistency of the CSP, a new bounded planning problem allowing more actions is generated and the process restarts.

In addition to this iterative deepening setting, LCP also supports a configuration that defines the size of the planning problem to solve: specifying for each action in the planning domain its maximal number of occurrences.
In this setting, LCP will generate a single bounded planning problem and attempt to find a solution for it, allowing a simple domain-dependent configuration of the planner.

\section{Case Study: The RoboCup Logistics League}
\label{sec:rcll}

\begin{figure}[t]
	\begin{center}
		\includegraphics[width=.7\textwidth]{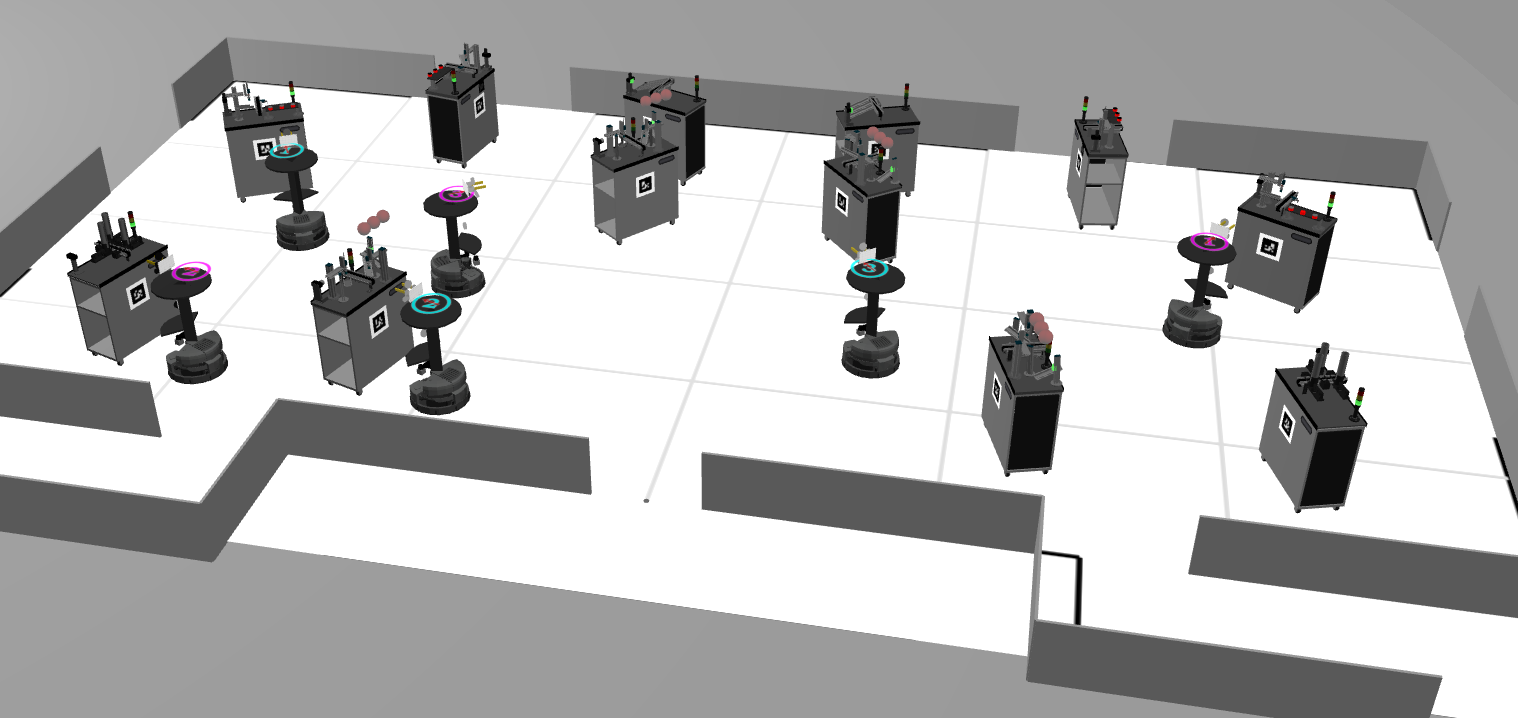}
	\end{center}
	\vspace{-4mm}
	\caption{Simulated RCLL factory environment~\cite{zwilling2014simulation}.}
	\label{fig:simulation}
\end{figure}

The RoboCup Logistics League (RCLL) provides a simplified smart factory scenario where two teams of three autonomous robots each
compete to handle the logistics of materials to accommodate orders known only at run-time. Competitions take place yearly using a real robotic setup, however, for our experiments we made use of the simulated environment (Fig.~\ref{fig:simulation}) developed for the Planning and Execution Competition for Logistics Robots in Simulation\footnote{\url{http://www.robocup-logistics.org/sim-comp}}~\cite{LogRobComp2016}.

\begin{figure}[t]
	\begin{center}
		\includegraphics[width=.95\textwidth]{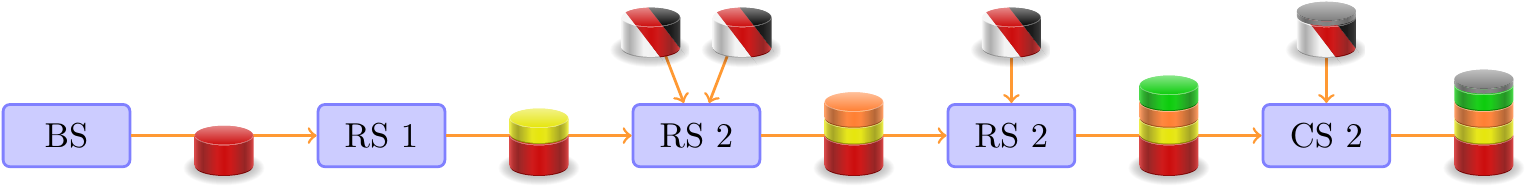}
	\end{center}
	\caption{Example of order configuration for the competition~\cite{PlanningInRCLL,RCLL-Rulebook-2017}. The order here depicted consists of a red base, three colored rings and a gray cap. Notice that orange and green rings require additional tokens (2 and 1 respectively)  in order to be mounted. }
	\label{fig:config}
\end{figure}

Products to be assembled have different complexities and usually
require a base, mounting $0$ to $3$ rings, and a cap as a finishing
touch. Bases are available in three different colors, four colors are admissible for rings and
two for caps, leading to about 250 different possible combinations. Each order defines which colors are to be used, together with an ordering.
An example of a possible configuration is shown in Fig.~\ref{fig:config}.

Several machines are scattered
around the factory shop floor (positions are different in each scenarios and announced to the robots at runtime). Each machine completes a particular production step such as providing bases, mounting colored rings or caps.
There are four types of machines:
\begin{itemize}
	\item Base Station (BS): acts as dispenser of base elements (one per team).
	\item Cap Station (CS): mounts a cap as the final step in production
	on an intermediate product. The CS stores at most one cap at a
	time (empty initially). To prefill the CS, a base element with a
	cap must be taken from a shelf in the arena and fed to the
	machine; the cap is then unmounted and buffered. The cap can then
	be mounted on the next intermediate product taken to the
	machine (two CS per team).
	\item Ring Station (RS): mounts one colored ring (of specific color)
	onto an intermediate product. Some ring colors require
	additional tokens: robots will have to feed a RS with a specified
	number of bases before the required color can be mounted (two RS
	per team).
	\item Delivery Station (DS): accepts finished products (one per
	team).
\end{itemize}

The objective for a team of autonomous robots is to transport intermediate
products between processing machines and optimize a multistage production cycle of different product variants until delivery of final products.

Orders that denote the products which must be assembled are posted at
run-time by an automated referee box and come with a delivery time
window, therefore introducing a temporal component that requires quick
planning and scheduling.

\section{Experimental evaluation}
\label{sec:experiments}
\def \omtsat {\omtplan-Sat\xspace}
\def \omtopt {\omtplan-Opt\xspace}
\def \macrosat {\omtplan-Macros-Sat\xspace}
\def \macroopt {\omtplan-Macros-Opt\xspace}
\def \lcpgen {LCP-Gen\xspace}
\def \lcpspe {LCP-Spe\xspace}
\def \optic  {Optic\xspace}
\def \smtplan {SMTPlan+\xspace}

\def \solved {Solved}
\def \runtime {Runtime (s)}
\begin{table*}
    \centering
    \begin{tabular}{|l|c|c|c|c|c|c|} \hline 
        & \multicolumn{2}{c|}{C0 -- R1} & \multicolumn{2}{c|}{C0 -- R2} & \multicolumn{2}{c|}{C0 -- R3} \\ \hline
        Planner & \solved & \runtime & \solved & \runtime & \solved & \runtime \\ \hline\hline
        \omtsat & \textbf{20} & 0.98 & \textbf{20} & 2.52 & \textbf{20} & 2.73 \\\hline
        \omtopt & \textbf{20} & 15.22 & 17 & 42.18 & 4 & 54.26 \\\hline\hline
        \macrosat & \textbf{20} & \textbf{0.13} & \textbf{20} & \textbf{0.14} & \textbf{20} & \textbf{0.14} \\\hline
        \macroopt & \textbf{20} & 0.40 & \textbf{20} & 0.53 & \textbf{20} & 0.60 \\\hline\hline
        \lcpgen & \textbf{20} & 7.25 & \textbf{20} & 9.72 & \textbf{20} & 11.52 \\\hline
        \lcpspe & \textbf{20} & 1.61 & \textbf{20} & 1.93 & \textbf{20} & 2.74 \\\hline\hline
        \optic & 13 & 24.39 & 2 & 57.60 & 0 & -- \\\hline
        \smtplan & 0 & -- & 0 & -- & 0 & -- \\ \hline 
    \end{tabular}\\[5pt]
    \caption{Results for the C0 configuration.
     For problems with 1 to 3 robots (R1--R3), it indicates for each planner the number of problems solved and the average runtime when a plan was found before the timeout of 60 seconds.}
     \label{tab:C0}
\end{table*}

\begin{table*}
    \centering
    \begin{tabular}{|l|c|c|c|c|c|c|} \hline
        & \multicolumn{2}{c|}{C1 -- R1} & \multicolumn{2}{c|}{C1 -- R2} & \multicolumn{2}{c|}{C1 -- R3} \\ \hline
        Planner & \solved & \runtime & \solved & \runtime & \solved & \runtime \\ \hline\hline
        \omtsat & \textbf{20} & 11.55 & 16 & 10.35 & 18 & 24.11 \\\hline
        \omtopt & 0 & -- & 0 & -- & 0 & -- \\\hline\hline
        \macrosat & \textbf{20} & \textbf{0.37} & \textbf{20} & \textbf{0.43} & \textbf{20} & \textbf{0.60} \\\hline
        \macroopt & \textbf{20} & 1.82 & \textbf{20} & 9.57 & 17 & 6.48 \\\hline\hline
        \lcpgen & 2 & 30.07 & 1 & 56.12 & 1 & 48.67 \\\hline
        \lcpspe & \textbf{20} & 7.03 & 15 & 10.59 & 12 & 12.45 \\\hline\hline
        \optic & 12 & 23.46 & 0 & -- & 0 & -- \\\hline
        \smtplan & 0 & -- & 0 & -- & 0 & -- \\\hline
    \end{tabular}\\[5pt]
    \caption{Results for the C1 configuration.
     For problems with 1 to 3 robots (R1--R3), it indicates for each planner the number of problems solved and the average runtime when a plan was found before the timeout of 60 seconds.}
     \label{tab:C1}
\end{table*}

\subsection{Experimental Setup}

We evaluate our various planning components against the RCLL set of benchmarks.
Unlike the setting for the RCLL competition that features two competing teams of robots, we use a single team which is closer to the real-world setting in which robots in the same factory are expected to collaborate in the production steps.

The PDDL domain is taken, unmodified, from the referee box of RCLL competition.%
\footnote{\url{https://github.com/timn/ros-rcll_ros/tree/master/pddl}}
Problems are generated for different game settings and vary on the number of robots available (from 1 to 3 robots). The complexity of the product to manufacture varies between complexities C0 and C1, respectively corresponding to no-ring and one-ring configurations of the final product.

\subsection{Tested planners}

Our special purpose planning component is evaluated in its two version, both declined in two subversions depending on whether they seek feasible or optimal plans.
\begin{itemize}
    \item \omtsat \& \omtopt represent the basic RCLL-specific planning component, respectively configured to seek feasible and optimal plans.
    While this encoding is domain specific, it closely mimics the actions available in the executive system.
    \item \macrosat \& \macroopt represent the RCLL-specific planning component with macro actions. They are respectively configured to seek feasible and optimal plans.
    Unlike, \omtsat/\omtopt, the use of macro actions allows for much smaller encoding at the expense of divergence with the underlying execution system and more complexity in its development.
\end{itemize}

The LCP component is evaluated in two configurations:
\begin{itemize}
    \item \lcpgen is our domain-independent planning component running with default options.
    \item \lcpspe is our domain-independent planning component with a configuration tailored to the RCLL domains.
    Namely, the bounded planning problem is restricted to actions that might be useful in the RCLL domain.
    This configuration is purely external and does not require touching the internals of the planner.
\end{itemize}  

Finally, we compare \ourtool{} to two state-of-the-art temporal planners:
\begin{itemize}
    \item \optic \cite{Benton2012} is a forward-search heuristic planner evolved from the POPF \cite{Coles2010} planner which was a runner-up the penultimate International Planning Competition (IPC).
    \optic is a forward search heuristic planner that uses a temporal extension of the $h^{FF}$ heuristic.
    TFD and YAHSP, the other top competitors of the latest IPCs, were not considered because \emph{(i)} they are not complete with respect to the semantics of PDDL~2.1 \cite{cushing2007}, and \emph{(ii)} they do not support the PDDL encoding of the RCLL domain that mixes instantaneous and durative actions.
    \item \smtplan is a recent SMT-based planner for the PDDL+ language \cite{smtplan}. 
    \smtplan is more expressive than the other planners considered as it supports the full range of PDDL+ features, including continuous processes inducing non-linear changes over numeric state variables.
    It uses a state-oriented SMT encoding, over plans of increasing length.
\end{itemize}

All SMT-based planners use Z3 \cite{z3} in version 4.6.3 as an off-the-shelf solver.

\subsection{Results}

All benchmarks were run on an Intel i7-3770 @ 3.40GHz with a timeout of 60 seconds.
The timeout is low compared to the usual setting of 30 minutes of the International Planning Competition but is more adequate to an online planning setting such as the one of RCLL.

\subsubsection{C0 configuration}
Results for the C0 configuration of the RCLL problem are given in Table~\ref{tab:C0}.  
For a each planner, the table provides the number of problems solved and the average runtime for successful runs.

For the C0 configuration, all components of \ourtool are able to solve all 60 problems. 
The only exception is the base configuration of the domain-dependent approach that fails to prove optimality for a number of problems (\omtopt).
Runtimes are largely dominated by the domain-specific encoding with macro actions: runtimes are well below a second for both feasibility and optimality (\macrosat/\macroopt).
Given their generality, both versions of LCP provide good performance solving all problems in handful of seconds.
Notably, performance of the configured version of LCP (\lcpspe) is on par with that of \omtsat. 

\optic has overall poor performance on those domains, solving only 13 problems involving 1 robot and 2 problems involving 2 robots.
\smtplan fails to solve any attempted problem.

\subsubsection{C1 configuration}

Problems for configuration C1, where an additional ring must be mounted on the product, show more differentiated results (Table~\ref{tab:C1}).
The benchmark is still dominated by the domain specific macro-encoding, that only fails to prove the optimality of three of the most difficult problems.

\omtsat and \lcpspe continue to show comparable performance both in runtime and number of problems solved in R1 and R2.
Both \omtopt and \lcpgen show their limits, solving respectively none and 4 problems.

\optic almost maintains its performance from the C0 configuration, solving 12 of the 20 problems involving 1 robot.

\section{Discussion and Conclusion}
\label{sec:conclusion}

In this paper we have presented and evaluated \ourtool, a planner that leverages the latest SMT technology for solving typical problems that arise in smart factories.
Evaluation of the domain-specific and domain-independent components of \ourtool on the RCLL benchmarks highlight different trade-offs.

The most involved domain-specific solver, shows unchallenged performance on the RCLL benchmarks highlighting the work that remains to be done on fully automated task planners.
Of course the development of such a domain-specific planner induces many difficulties as it is more error prone and time consuming.
Perhaps most importantly, adapting it to new operational constraints is challenging as the process must be restarted to account for violated assumptions.

On the other hand, fully domain-independent planners provide a great promise of reusability and adaptability to a wide variety of contexts.
The development of LCP as a part of \ourtool is meant to leverage those benefits.
While the gap with domain specific solvers remains important, experiments show that LCP does reduce this gap with respect to existing domain-independent planners.
In its current state, it would stand as a valid solution when provided with minimal configuration.

The performance gap indicates two directions for future work on LCP.
First there is the never-ending quest for performance improvement, the performance of \omtplan defining a challenging target to reach.
Second, we believe the performance gain from injecting domain-specific knowledge (being in specific solvers or in the configuration of LCP) raises the question of how to inject such knowledge in a principled and solver-independent way.
The optional specification of Hierarchical Task Networks (HTN) in the ANML language provide a good opportunity to do this in an incremental and non-intrusive manner.
The ANML subset for HTN is currently not supported by LCP and will be the subject of future work.

\section*{Acknowledgements}
The authors would like to thank Igor Bongartz for his help
with the implementation and validation of \omtplan.

\bibliographystyle{unsrt}

\begin{thebibliography}{10}

\bibitem{zhong2017intelligent}
Ray~Y Zhong, Xun Xu, Eberhard Klotz, and Stephen~T Newman.
\newblock Intelligent manufacturing in the context of industry 4.0: A review.
\newblock {\em Engineering}, 3(5), 2017.

\bibitem{dblp:series/faia/barrettsst09}
Clark~W. Barrett, Roberto Sebastiani, Sanjit~A. Seshia, and Cesare Tinelli.
\newblock {Satisfiability Modulo Theories}.
\newblock In {\em Handbook of Satisfiability}. IOS Press, 2009.

\bibitem{dblp:conf/tacas/cimattifgss10}
Alessandro Cimatti, Anders Franz{\'{e}}n, Alberto Griggio, Roberto Sebastiani,
  and Cristian Stenico.
\newblock Satisfiability modulo the theory of costs: Foundations and
  applications.
\newblock In {\em Tools and Algorithms for the Construction and Analysis of
  Systems (TACAS)}, 2010.

\bibitem{z3}
Leonardo~Mendon{\c{c}}a de~Moura and Nikolaj Bj\o{}rner.
\newblock Z3: An efficient {SMT} solver.
\newblock In {\em Tools and Algorithms for the Construction and Analysis of
  Systems (TACAS)}, 2008.

\bibitem{dblp:conf/tacas/bjornerpf15}
Nikolaj Bj\o{}rner, Anh{-}Dung Phan, and Lars Fleckenstein.
\newblock {\(\nu\)}z - an optimizing {SMT} solver.
\newblock In {\em Tools and Algorithms for the Construction and Analysis of
  Systems (TACAS)}, 2015.

\bibitem{DBLP:conf/cav/CimattiG12}
Alessandro Cimatti and Alberto Griggio.
\newblock Software model checking via {IC3}.
\newblock In {\em International Conference on Computer Aided Verification
  ({CAV})}, 2012.

\bibitem{DBLP:conf/cade/TiwariGD15}
Ashish Tiwari, Adri{\`{a}} Gasc{\'{o}}n, and Bruno Dutertre.
\newblock Program synthesis using dual interpretation.
\newblock In {\em International Conference on Automated Deduction (CADE)},
  2015.

\bibitem{smtplan}
Michael Cashmore, Maria Fox, Derek Long, and Daniele Magazzeni.
\newblock A compilation of the full {PDDL+} language into {SMT}.
\newblock In {\em International Conference on Automated Planning and Scheduling
  (ICAPS)}, 2016.

\bibitem{bit-monnot-cp-2018}
Arthur Bit-Monnot.
\newblock {A Constraint-based Encoding for Domain-Independent Temporal
  Planning}.
\newblock In {\em International Conference on Principles and Practice of
  Constraint Programming (CP)}, 2018.

\bibitem{dblp:conf/iri/leofanteanlt17}
Francesco Leofante, Erika {\'{A}}brah{\'{a}}m, Tim Niemueller, Gerhard
  Lakemeyer, and Armando Tacchella.
\newblock On the synthesis of guaranteed-quality plans for robot fleets in
  logistics scenarios via optimization modulo theories.
\newblock In {\em IEEE International Conference on Information Reuse and
  Integration (IRI)}, 2017.

\bibitem{dblp:journals/isf/leofanteanlt18}
Francesco Leofante, Erika {\'{A}}brah{\'{a}}m, Tim Niemueller, Gerhard
  Lakemeyer, and Armando Tacchella.
\newblock Integrated synthesis and execution of optimal plans for multi-robot
  systems in logistics.
\newblock {\em (under review)}, 2018.

\bibitem{PlanningInRCLL}
Tim Niemueller, Gerhard Lakemeyer, and Alexander Ferrein.
\newblock The robocup logistics league as a benchmark for planning in robotics.
\newblock In {\em ICAPS Workshop on Planning and Robotics (PlanRob)}, 2015.

\bibitem{bit-monnot2016}
Arthur Bit-Monnot.
\newblock {\em {Temporal and Hierarchical Models for Planning and Acting in
  Robotics}}.
\newblock PhD thesis, Universit{\'{e}} de Toulouse, 2016.

\bibitem{DBLP:conf/ecai/KautzS92}
Henry~A. Kautz and Bart Selman.
\newblock Planning as satisfiability.
\newblock In {\em European Conference on Artificial Intelligence (ECAI)}, 1992.

\bibitem{DBLP:conf/tacas/CimattiGSS13}
Alessandro Cimatti, Alberto Griggio, Bastiaan~Joost Schaafsma, and Roberto
  Sebastiani.
\newblock {The MathSAT5 SMT solver}.
\newblock In {\em Tools and Algorithms for the Construction and Analysis of
  Systems (TACAS)}, 2013.

\bibitem{DBLP:conf/sat/CorziliusKJSA15}
Florian Corzilius, Gereon Kremer, Sebastian Junges, Stefan Schupp, and Erika
  {\'{A}}brah{\'{a}}m.
\newblock {SMT-RAT:} an open source {C++} toolbox for strategic and parallel
  {SMT} solving.
\newblock In {\em International Conference on Theory and Applications of
  Satisfiability Testing (SAT)}, 2015.

\bibitem{DBLP:journals/tocl/SebastianiT15}
Roberto Sebastiani and Silvia Tomasi.
\newblock Optimization modulo theories with linear rational costs.
\newblock {\em {ACM} Transactions on Computational Logic}, 16(2), 2015.

\bibitem{DBLP:conf/cav/SebastianiT15}
Roberto Sebastiani and Patrick Trentin.
\newblock {OptiMathSAT}: A tool for optimization modulo theories.
\newblock In {\em International Conference on Computer Aided Verification},
  2015.

\bibitem{DBLP:conf/ictai/HoweyLF04}
Richard Howey, Derek Long, and Maria Fox.
\newblock {VAL: Automatic Plan Validation, Continuous Effects and Mixed
  Initiative Planning Using PDDL}.
\newblock In {\em IEEE International Conference on Tools with Artificial
  Intelligence (ICTAI)}, 2004.

\bibitem{leofanteanlt2018}
Francesco Leofante, Erika {\'{A}}brah{\'{a}}m, Tim Niemueller, Gerhard
  Lakemeyer, and Armando Tacchella.
\newblock Integrated synthesis and execution of optimal plans for multi-robot
  systems in logistics.
\newblock {\em Information Systems Frontiers}, 2018.

\bibitem{Fox2003}
Maria Fox and Derek Long.
\newblock {PDDL2.1: An Extension to PDDL for Expressing Temporal Planning
  Domains}.
\newblock {\em Journal of Artificial Intelligence Research (JAIR)}, 20, 2003.

\bibitem{anml2008}
David~E. Smith, Jeremy Frank, and William Cushing.
\newblock {The ANML Language}.
\newblock In {\em International Conference on Automated Planning and Scheduling
  (ICAPS)}, 2008.

\bibitem{ghallab2004}
Malik Ghallab, Dana~S. Nau, and Paolo Traverso.
\newblock {\em {Automated Planning: Theory and Practice}}.
\newblock Elsevier, 2004.

\bibitem{ghallab1994}
Malik Ghallab and Herv{\'{e}} Laruelle.
\newblock {Representation and Control in IxTeT, a Temporal Planner}.
\newblock In {\em International Conference on Artificial Intelligence Planning
  and Scheduling (AIPS)}, 1994.

\bibitem{laborie2008}
Philippe Laborie and Jer{\^{o}}me Rogerie.
\newblock {Reasoning with Conditional Time-Intervals}.
\newblock In {\em International Florida Artificial Intelligence Research
  Society Conference (FLAIRS)}, 2008.

\bibitem{dvorak-2014-ictai}
Filip Dvor{\'{a}}k, Roman Bart{\'{a}}k, Arthur Bit-Monnot, F{\'{e}}lix Ingrand,
  and Malik Ghallab.
\newblock {Planning and Acting with Temporal and Hierarchical Decomposition
  Models}.
\newblock In {\em IEEE International Conference on Tools with Artificial
  Intelligence (ICTAI)}, 2014.

\bibitem{laborie2009}
Philippe Laborie, J{\'{e}}r{\^{o}}me Rogerie, Paul Shaw, and Petr Vil{\'{\i}}m.
\newblock {Reasoning with Conditional Time-Intervals. Part II: An Algebraical
  Model for Resources.}
\newblock In {\em International Florida Artificial Intelligence Research
  Society Conference (FLAIRS)}, 2009.

\bibitem{zwilling2014simulation}
Frederik Zwilling, Tim Niemueller, and Gerhard Lakemeyer.
\newblock Simulation for the {RoboCup Logistics League} with real-world
  environment agency and multi-level abstraction.
\newblock In {\em Robot Soccer World Cup}. Springer, 2014.

\bibitem{LogRobComp2016}
Tim Niemueller, Erez Karpas, Tiago Vaquero, and Eric Timmons.
\newblock Planning competition for logistics robots in simulation.
\newblock In {\em ICAPS Workshop on Planning and Robotics (PlanRob)}, 2016.

\bibitem{RCLL-Rulebook-2017}
{RCLL Technical Committee}.
\newblock {RoboCup Logistics League} -- rules and regulations, 2017.

\bibitem{Benton2012}
J.~Benton, Amanda Coles, and Andrew Coles.
\newblock {Temporal Planning with Preferences and Time-Dependent Continuous
  Costs}.
\newblock In {\em International Conference on Automated Planning and Scheduling
  (ICAPS)}, 2012.

\bibitem{Coles2010}
Amanda Coles, Andrew Coles, Maria Fox, and Derek Long.
\newblock {Forward-Chaining Partial-Order Planning}.
\newblock In {\em International Conference on Automated Planning and Scheduling
  (ICAPS)}, 2010.

\bibitem{cushing2007}
William Cushing, Subbarao Kambhampati, Mausam, and Daniel~S. Weld.
\newblock {When is Temporal Planning Really Temporal?}
\newblock In {\em International Joint Conference on Artificial Intelligence
  (IJCAI)}, 2007.

\end{thebibliography}

\end{document}